# MedSceneBench: Evaluating Clinical Issue Discovery in Full-Length Medical Drama Episodes

Yilin Yang,[1] Tinaz Bhutta, M.D.,[2] Vishal Busa, M.D.,[2] Zhenghui Guo,[1] Nicholas C. Wlezien,[1] Armaan Amatya,[1] Omprakash Gnawali, Ph.D.,[1] Chengming Zhang, Ph.D.,[1] and Jivan Lamichhane, M.D.[2,†]
[1]University of Houston, Houston, TX
[2]SUNY Upstate Medical University, Syracuse, NY
[†]Corresponding author: Jivan Lamichhane (lamichhj@upstate.edu).

## Abstract

**BACKGROUND** Evaluating hospital care requires identifying clinical issues amid evolving patient conditions, clinical decisions, and team interactions. Current video description and question- answering benchmarks do not test whether models can independently recognize these issues across the entire episode of care.

**METHODS** We introduce MedSceneBench, a clinician-annotated benchmark containing 306 clips and 795 clinical issues from 151 medical drama episodes. Each issue was assigned to one of four categories: efficiency, ethical violations, breaches of standards of care, and patient outcomes. We tested 10 open-source models on their ability to identify clinical issues and to report the category, time interval, and explanation for each. We compared complete-episode input (Direct-Full) with Caption-Memory-20m, which inputs the videos to the models in 20-minute segments. Detection F1 measured category and time matching. BERTScore and a large language model judge measured agreement with clinician explanations.

**RESULTS** All 10 models completed Direct-Full, and five completed Caption-Memory-20m. The highest detection Macro-F1 was 0.58% with Direct-Full and 1.07% with Caption-Memory-20m. Across both settings, models matched 52 of 795 distinct annotated issues. The judge scored 50 of 53 matched explanations 0 or 1 on a 0–3 scale. With incomplete runs scored as zero, caption memory improved Macro-F1 for one of the 10 models.

**CONCLUSIONS** MedSceneBench tests clinical issue discovery, temporal localization, and explanation within hospital narratives. The evaluated models missed most annotated issues and often described a different issue even when the category and time matched.

**Dataset:** MedSceneBench on Hugging Face Dataset Repository

## Introduction

Hospital care depends on the actions and decisions of patients, clinicians, nurses, and other team members. Judging that care requires knowing who did what, what the patient needed at the time, whether required safety measures were followed, and what happened before and after each action. Lapses in care are common in real hospitals. Interns often spend most of their time away from patients [1], diagnostic tests are often overused [2], patients often do not fully understand what they consent to [3], patient information is shared outside the care team [4], and staff clean their hands less often than they report [5].

Studies of medical dramas have examined resuscitation outcomes [6], infection precautions [7], medication orders [8], drug choices [9], delirium [10], ethical decisions [11], and responses to social needs [12]. Others have compared how clinicians spend their time on screen and in real hospitals [13], documented unprofessional behavior [14], and assessed the quality of resuscitation [15] and survival after it [16]. Educators also use these scenes to teach patient safety [17,18]. Existing benchmarks test important parts of this problem. TVQA, Video-MME, and LongVideoBench evaluate question-guided video understanding, while MLVU also includes captioning and summarization [19-22]. Medical video benchmarks cover instructional understanding, surgical procedures, and clinically grounded questions [23-25]. Research on operating rooms, outpatient encounters, complete clinical procedures, and streaming alerts addresses workflow and safety [26-30]. Other benchmarks ask models to locate physical examinations [31], answer questions about procedures in MedHorizon [29], find and describe events in LongVALE [32], and link film events in NEST [33]. These benchmarks do not directly test whether a model can independently find varied clinical issues across a hospital narrative. Question-guided tasks can indicate which event to examine, and an accurate event description does not by itself establish whether the care is appropriate.

Benchmarks based on clinical text and electronic health records also show the value of defining tasks around the work clinicians perform. CORAL uses expert annotations to evaluate extraction of detailed oncology information from clinical narratives [34]. MedAgentBench evaluates tasks performed by language-model agents in an interactive medical-record environment [35]. These resources connect model outputs to explicit clinical references or verifiable task outcomes. They do not test whether models can independently find clinical issues throughout a complete video and locate the evidence for each issue.

Three connected challenges make independent discovery difficult. First, clinical judgment depends on context, including the patient's condition, authorization, supervision, and preceding decisions. An issue may also be an omission, such as a missed step or a delay. Second, edited videos move between patients and later return to earlier encounters. Models must connect the correct people and events across shots and avoid reporting one issue repeatedly. Third, later actions may depend on information from much earlier in the episode. Models must retain that context and locate the time interval supporting each issue.

We introduce MedSceneBench, a clinician-annotated benchmark for clinical issue discovery in complete medical drama episodes. These episodes show bedside care, emergencies, medical procedures, and interactions among clinical staff. Each model receives an episode and the same instructions, with no question that points to a specific event. Models must discover clinical issues, identify their categories and time intervals, and explain why they are issues. We compare 10 open-source models, including general-purpose models and a medical-domain model. We also give the annotated clips directly to 11 open-source models and 2 closed-source models (Claude Opus 5.5 and GPT-6.1 Sol) to test whether they can classify and explain the issue categories when the relevant scene is provided.

# Methods

## Dataset Sources and Characteristics

MedSceneBench uses episodes from six medical television series. We use complete episodes containing clinician-annotated clips so that models must find issues within their surrounding context.

The final annotated dataset contains 306 clips from 151 episodes, with 795 clinical issues. Table 1 groups the annotated episodes and clips by series title. The 151 annotated episodes used for evaluation have a total running time of 110.67 hours. Episodes ran 40-57 minutes (median 43.75 minutes).

| Table 1. Dataset coverage by series. | | |
|---|---|---|
| **Series** | **Episodes** | **Annotated clips** |
| The Pitt | 9 | 12 |
| House M.D. | 20 | 53 |
| Pulse | 6 | 10 |
| Grey's Anatomy | 54 | 90 |
| Code Black | 5 | 11 |
| The Resident | 57 | 130 |
| Total | 151 | 306 |

## Annotation Format

Each record describes a clinician-selected clip, with identifiers, start and end times, a clinical summary, and four issue category fields (Figure 1):

- **Efficiency:** an inefficiency involving time, workflow, team or resource use, or an unnecessarily slow or burdensome approach.
- **Ethical Violations:** an ethical concern involving consent, autonomy, privacy, coercion, professional responsibility, or respect for the patient.
- **Breach of Standards of Care:** a departure from an appropriate standard of care, required safeguard, supervision practice, or clinically appropriate procedure.
- **Patient Outcome:** an actual adverse patient outcome or a clear potential risk of patient harm arising from the care shown.

The clinical team defined these categories, which match aspects of care that earlier studies of medical dramas examined [13-16].

Each non-empty category field contains a clinician-written description and defines one annotated issue, even when it mentions several actions. Issues from the same clip share its time interval but have separate categories and explanations. The annotated issue is the unit of evaluation. A blank category field indicates no annotated issue of that type within the clip; areas outside annotated clips have no reference labels. Supplementary Section S2 describes the interval conventions. Supplementary Table S3 gives an unabridged annotation example.

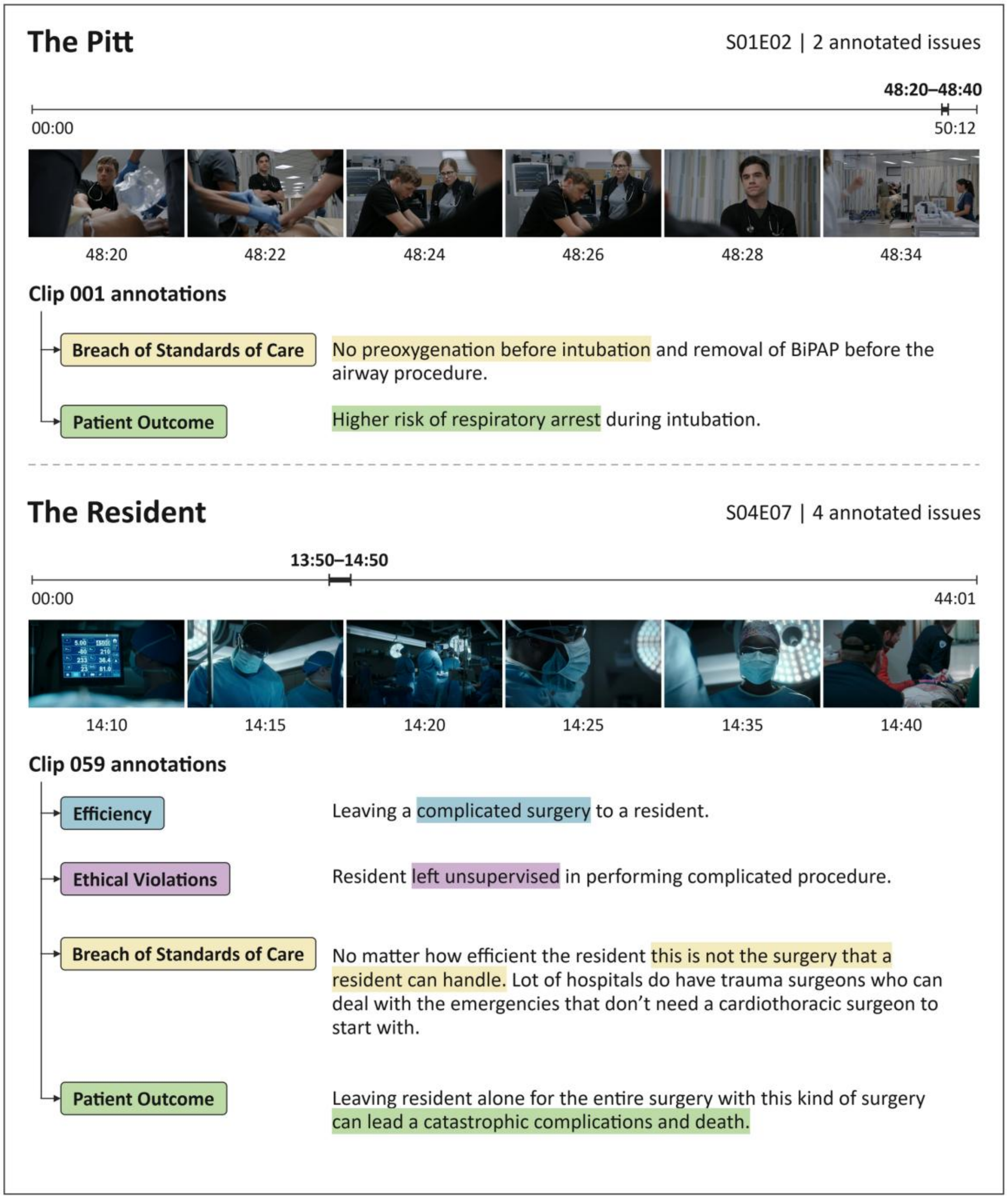

***Figure 1.*** *Examples of clinical issue annotations in MedSceneBench. The frames come from The Pitt (S01E02) and The Resident (S04E07). Their timestamps refer to the full episodes. One annotated clip can contain several annotated issues. The timelines show where each clip appears in the full episode. Issue descriptions are reproduced verbatim from the annotation records. Highlighting was added for this figure to link phrases to issue categories.*

*Source: MedSceneBench annotation records; video frames from The Pitt (S01E02) and The Resident (S04E07).*

## Clinical Annotation Process

Two faculty clinicians at SUNY Upstate Medical University, both board-certified in internal medicine, independently reviewed clinical care and medical decisions shown. They judged possible problems using only the events and information from the episode. The review covered patient presentations, diagnostic reasoning and tests, treatment choices, procedural techniques, and adherence to clinical standards. Reviewers recorded each relevant clip's start and end times, summarized the clinical situation, and described each issue and why it was a concern. They assigned all applicable categories and refined the annotations while preserving clinical context and reasoning.

## Quality Control for Annotations

Two quality improvement and patient safety officers, board-certified internal medicine faculty, each reviewed the annotations for 15 randomly sampled clips (30 clips in all, 9.8% of 306). Of the 73 annotated issues they assessed, they agreed that 55 (75%) were present. Agreement was highest for Breach of Standards of Care (26 of 28 issues) and lowest for Efficiency (9 of 19 issues). They rated 44 of these 55 descriptions as accurate. Agreement was 18 of 34 issues for one officer and 37 of 39 for the other. Their clips came mostly from different annotation batches. The officers also found 9 issues in categories that the annotators had left blank. As a result, some model predictions counted as false positives may be real but unannotated issues.

## Evaluation Protocol and Scoring

Figure 2 shows how we generate episode-level predictions and match them to clinician annotations.

**Step 1: Generate episode-level predictions.** We feed the model a video and a prompt that defines the four issue categories and specifies the required output format. **Direct-Full** provides the complete episode and its duration. **Caption-Memory-20m** provides the current video segment, its episode-relative time range, and the model's captions from earlier segments. The model must identify clinical issues without event-specific questions, clinician summaries, annotated clips, or reference intervals.

The model, for each issue, returns its category, episode-relative start and end times, and explanation. Valid predictions require a recognized category, a non-empty explanation, and a positive-duration interval within the video provided in that request. The prompt asks the model to reply with "No issues found." if no issues are identified. Absence of these positive or negative valid response is identified as failure and contributes no matches. If a response contains both valid and malformed issue blocks, valid blocks remain eligible for matching. Supplementary Section S4 provides the output format and prompts.

Direct-Full uses one request and returns one issue list. Caption-Memory-20m processes consecutive, non-overlapping 20-minute segments. For each segment, the model returns a summary (caption) and an issue list. All earlier summaries, but not predicted issues, accompany the next segment. We combine all segment issue lists into one episode-level output. A complete run requires a response for each of the 151 episodes under Direct-Full or each of the 453 segments under Caption-Memory-20m. If a run is incomplete or all its responses are unusable, we assign zero to all three metrics.

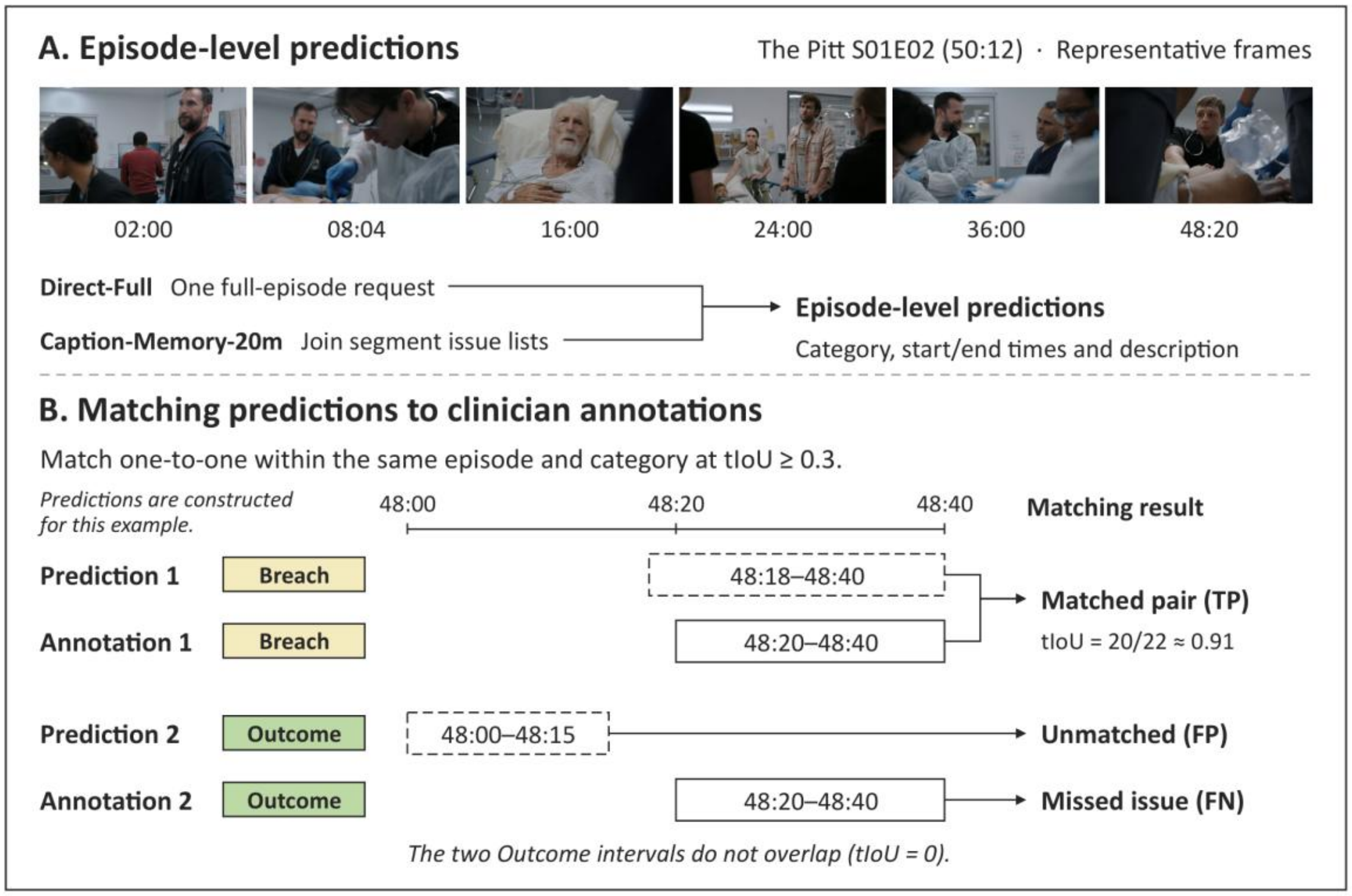

***Figure 2.*** *Episode-level prediction and matching in MedSceneBench. Panel A shows how each protocol generates episode-level predictions. Panel B compares two illustrative predictions with clinician annotations from The Pitt (S01E02). Breach and Outcome denote Breach of Standards of Care and Patient Outcome, respectively.*

**Step 2: Match predictions to annotated issues.** Temporal intersection over union (tIoU) measures overlap between predicted and annotated intervals [36,37]. A predicted issue $p$ can match an annotated issue $g$ only within the same episode and category, with tIoU at least $\tau = 0.3$. One-to-one assignment prioritizes the number of pairs meeting the threshold and then their temporal overlap. Each prediction and annotation can enter at most one match. The accepted pairs form $\mathcal{M}_\tau$. Supplementary Section S3 gives matching details and worked scoring examples.

**Step 3: Aggregate detection and explanation scores.** Matched predictions are true positives (TP); unmatched predictions with a recognized category are false positives (FP) for that category, and unmatched annotations are false negatives (FN). We sum these counts across evaluated episodes for each category $c$, then calculate F1 and average the four category scores with equal weight:

$$F1_c = \frac{2TP_c}{2TP_c + FP_c + FN_c}, \qquad \text{Macro-F1} = \frac{1}{4}\sum_{c=1}^{4} F\,1_c.$$

Macro-F1@0.3 specifies the tIoU threshold.

For each matched pair, we compare explanations using BERTScore-F1, $F_B(p, g)$, and a large language model (LLM) judge score, $J(p, g)$. BERTScore compares words in context [38]. The judge compares the model's explanation with the clinician's annotation, using the issue category and its definition. It assesses agreement with the clinician's issue description on a 0–3 scale adapted from MMBench-Video [39].

Both text metrics use end-to-end (E2E) averaging over all annotated issues. Let $\mathcal{M}_\tau$ be the matched pairs from Step 2, pooled across evaluated episodes and categories, and let $N_G$ be the total number of annotated issues in that evaluation scope. Then

$$B_{\text{E2E}} = \frac{1}{N_G} \sum_{(g,p)\in\mathcal{M}_\tau} F_B\,(p,g), \qquad L_{\text{E2E}} = \frac{1}{N_G} \sum_{(g,p)\in\mathcal{M}_\tau} \frac{J(p,g)}{3}.$$

Judge scores are normalized to 0–1 before averaging. The denominator includes all annotated issues, with missed annotations contributing zero; it is not the number of matched pairs. In the result tables, Macro-F1, BERTScore-E2E, and LLMScore-E2E are multiplied by 100 and rounded to two decimal places. Pairwise BERTScore-F1 and LLM judge scores are reported on their original scales.

Category and time matching locate the annotated context; the text scores assess whether the prediction expresses the same clinical issue.

### Annotated-Clip Classification

To separate recognizing an issue from locating it, we also gave models each clinician-selected clip directly. For each clip, the model decided whether each of the four categories was present. Four prompt formats requested the labels in one response (P1), after a description of the clip (P2), together with issue descriptions (P3), or one category at a time (P4). Supplementary Section S6 gives the prompts. A blank category field counts as no issue, giving 1,224 clip-category labels, of which 795 (65%) are positive. We report Macro-F1 across categories and compare it with a random baseline that answers yes for each category with probability 0.5. This evaluation adds Baichuan-Omni-1.5 [40], MiniCPM-o 4.5 [41], Claude Opus 5.5, and GPT-6.1 Sol, and excludes Qwen3-VL-32B. The closed-source models received frames at 1 frame per second, up to 512 per clip, without audio.

## Results

### Experimental Setup

For the Direct-Full and Caption-Memory-20m evaluations, we report results for 10 open-source models that are listed on Table 2. These models include the state-of-the-art vision language models, including one that was trained on medical videos. Direct-Full and Caption-Memory-20m differ in context, sampling, and request budgets and are compared as complete inference strategies. Further details of the experiments are in the supplementary document.

### Detection across Complete Episodes

Table 2 reports results against 795 clinician-annotated issues from 306 clips in 151 episodes. All 10 models completed Direct-Full. Five completed all 453 Caption-Memory-20m segment calls. The five incomplete runs received zero on all three metrics. Qwen3.8-27B returned 360 segments and Kimi-VL returned 51. Qwen3-Omni, InternVL3.5-8B, and VideoLLaMA3-7B produced no formal segment outputs.

Four of 10 models matched at least one annotation under Direct-Full, yielding 40 matched pairs. Qwen3.8-27B had the highest Macro-F1 (0.58%). Qwen3-VL-32B matched the most annotations (16 of 795) and had the highest LLMScore-E2E (0.46 on the 0–100 scale). Its BERTScore-E2E of 1.73 and GLM-4.6V-Flash's score of 1.28 both appear as 1.13 under the table's display cap. Under Caption-Memory-20m, two of the five models that completed the protocol matched an annotation. GLM-4.6V-Flash matched 12 and had the highest Macro-F1 (1.07%). Qwen3-VL-32B matched one.

**Table 2. Complete-episode performance with Direct-Full and Caption-Memory-20m.**

| Model | Direct-Full | | | Caption-Memory-20m | | |
|---|---|---|---|---|---|---|
| | Macro-F1 @0.3 | BERTScore -E2E | LLMScore -E2E | Macro-F1 @0.3 | BERTScore -E2E | LLMScore -E2E |
| Qwen3-VL-8B [42] | 0.10 | 0.32 | 0.08 | 0.00 | 0.00 | 0.00 |
| Qwen3.8-27B [43] | 0.58 | 0.96 | 0.25 | 0.00 | 0.00 | 0.00 |
| Qwen3-Omni [49] | 0.00 | 0.00 | 0.00 | 0.00 | 0.00 | 0.00 |
| MiniCPM-V 4.5 [44] | 0.00 | 0.00 | 0.00 | 0.00 | 0.00 | 0.00 |
| GLM-4.6V-Flash [45] | 0.46 | 1.13 | 0.29 | 1.07 | 1.13 | 0.25 |
| Kimi-VL [46] | 0.00 | 0.00 | 0.00 | 0.00 | 0.00 | 0.00 |
| InternVL3.5-8B [47] | 0.00 | 0.00 | 0.00 | 0.00 | 0.00 | 0.00 |
| VideoLLaMA3-7B [48] | 0.00 | 0.00 | 0.00 | 0.00 | 0.00 | 0.00 |
| SurgVidLM Stage 2 [24] | 0.00 | 0.00 | 0.00 | 0.00 | 0.00 | 0.00 |
| Qwen3-VL-32B [42] | 0.31 | 1.13 | 0.46 | 0.14 | 0.10 | 0.04 |

Direct-Full uses one request per episode. Caption-Memory-20m combines predictions from consecutive 20-minute segments on the episode timeline before scoring. A match requires a temporal intersection over union of at least 0.3 within the same episode and category. Scores are multiplied by 100 and rounded to two decimal places. BERTScore-E2E values above 1.13 are displayed as 1.13. Under Caption-Memory-20m, Qwen3.8-27B, Qwen3-Omni, Kimi-VL, InternVL3.5-8B, and VideoLLaMA3-7B received 0.00 on all three metrics because their runs were incomplete.

Six models had no Direct-Full matches. Three reported no issues, and the other three reported between 330 and 4,039 predictions each.

We tested Caption-Memory-20m because a long visual window might make it difficult to locate specific clinical issues. The protocol shortens the current visual input to consecutive 20-minute segments and retains earlier context through captions. Performance does not improve consistently, suggesting that shorter visual windows with text memory are not enough to resolve the difficulty of identifying and locating issues, despite explicit category definitions. Text memory may also omit visual details needed to judge a clinical event when earlier video is reduced to captions. Because the protocol also changes sampling and request budgets, the comparison does not isolate the effect of window length or rule out its contribution.

## Categories and Clinical Explanations

The 53 matched pairs across both protocols involved 52 distinct annotated issues. There were 33 Breach of Standards of Care matches, 11 Patient Outcome matches, 5 Efficiency matches, and 4 Ethical Violations matches. All four Ethical Violations matches came from Direct-Full. The judge assigned 0 to 23 of the 53 explanations, 1 to 27, and 2 to three. None received 3, the score for agreement on the core issue and its important qualifications.

For example, Qwen3-VL-8B reports an Efficiency issue for The Pitt, season 1, episode 8. Its interval matches an annotation about unrealistic timing of laboratory results. The prediction instead describes a nurse obtaining supplies from a slow automated cabinet. The pair has tIoU 0.425 and BERTScore-F1 0.871, but the judge assigns a score of 1 because the descriptions refer to different workflow problems. So a prediction can have the right category and time but still describe a different problem.

## Effect of the Inference Protocol

Across all 10 models, Caption-Memory-20m increased Macro-F1 for one, decreased it for three, and left six at zero. GLM-4.6V-Flash increased from 0.46% to 1.07%. Qwen3-VL-8B fell from 0.10% to zero and Qwen3-VL-32B from 0.31% to 0.14%. Qwen3.8-27B scored 0.58% under Direct-Full and received zero for its incomplete Caption-Memory-20m run. MiniCPM-V 4.5 and SurgVidLM Stage 2 had no matches in either setting.

GLM reported 3,289 issues with Direct-Full and 1,279 with Caption-Memory-20m. It matched 12 annotations in each setting, with one annotation shared across settings. False positives with a recognized category fell from 2,800 to 1,227. For Breach of Standards of Care, the largest such change, false positives fell from 2,335 to 512 while the number of matches changed from eight to seven. GLM's BERTScore-E2E was 1.28 for Direct-Full and 1.27 for Caption-Memory-20m before the display cap of 1.13. Its LLMScore-E2E fell from 0.29 to 0.25.

## Category Classification on Annotated Clips

Table 3 reports category classification when models receive the annotated clip. Only 4 of the 52 model and prompt combinations exceeded the random baseline of 54.87, and none of them used a closed-source model. The highest Macro-F1 was 71.71 (Baichuan-Omni-1.5, P3), and the closed-source models reached at most 41.40 (Claude Opus 5.5, P4) and 37.70 (GPT-6.1 Sol, P1). Scores changed widely with the prompt format; Baichuan-Omni-1.5 ranged from 17.09 under P2 to 71.71 under P3. Because 65% of the labels are positive, a model that answers yes more often obtains a higher Macro-F1 without distinguishing which categories apply. Supplementary Table S8 reports BERTScore-E2E and LLMScore-E2E for the issue descriptions under P3 and P4.

**Table 3. Category classification on annotated clips.**

| Model | P1 | P2 | P3 | P4 |
|---|---|---|---|---|
| Claude Opus 5.5 | 40.90 | 27.64 | 40.00 | 41.40 |
| GPT-6.1 Sol | 37.70 | 23.51 | 36.19 | 34.77 |
| Baichuan-Omni-1.5 [40] | 41.95 | 17.09 | 71.71 | 59.47 |
| Qwen3-Omni-30B-A3B [49] | 43.79 | 29.23 | 51.76 | 60.95 |
| MiniCPM-o 4.5 [41] | 56.50 | 46.96 | 48.88 | 52.92 |
| Qwen3-VL-8B [42] | 20.41 | 17.79 | 29.44 | 20.24 |
| VideoLLaMA3-7B [48] | 16.53 | 8.67 | 36.13 | 22.96 |
| SurgVidLM Stage 2 [24] | 35.78 | 14.32 | 39.11 | 41.29 |
| Qwen3.8-27B [43] | 17.24 | 21.02 | 21.39 | 30.79 |

| Table 3. Category classification on annotated clips. | | | | |
|---|---|---|---|---|
| **Model** | **P1** | **P2** | **P3** | **P4** |
| MiniCPM-V 4.5 [44] | 10.51 | 9.24 | 15.54 | 29.71 |
| GLM-4.6V-Flash [45] | 11.24 | 11.39 | 15.29 | 20.65 |
| Kimi-VL-A3B [46] | 19.91 | 13.25 | 16.13 | 7.50 |
| InternVL3.5-8B [47] | 3.42 | 3.23 | 11.32 | 9.78 |
| Random baseline (50% yes) | 54.87 | 54.87 | 54.87 | 54.87 |

Each model receives one clinician-selected clip and decides whether each of the four issue categories is present. P1–P4 denote the four prompt formats. Macro-F1 averages the four category F1 scores with equal weight and is shown as a percentage. The random baseline answers yes for each category with probability 0.5 and is shown as its expected Macro-F1. The closed-source models received video frames without audio.

Source: MedSceneBench annotated-clip evaluation.

# Discussion

MedSceneBench evaluates whether models can discover clinical issues throughout a hospital narrative and explain why the care shown is problematic. The results reveal difficulties at two stages. First, across all completed runs, the models matched the category and time interval of only 52 of the 795 annotated issues. Second, the judge found that only three of the 53 matched explanations identified the core clinical concern, and all three had an important omission, ambiguity, or minor factual error. These findings connect to issue discovery, temporal evidence, and clinical judgment as distinct parts of hospital-video understanding.

In this discovery setting, the model must identify which events constitute clinical issues and when they occur within a complete episode. The low scores show how demanding this combined task is and motivate progress in clinical context tracking and evidence-based explanations.

The distinction between describing an action and assessing care is central to this task. A model may describe what happened without identifying the clinical issue noted by the clinician. The relevant issue may depend on a missing safeguard, a preceding patient state, a professional obligation, or a delay. A useful explanation must connect the observed event to that context. Evaluating descriptions only for fluent language or familiar medical terms would miss this requirement.

The discrepancy between BERTScore and the judge scores illustrates why both detection and explanation metrics are needed. Two descriptions can share a category, a time interval, and common clinical vocabulary while referring to different problems. Scoring matched descriptions separately makes this difference visible, while E2E averaging also accounts for annotations that the model misses entirely.

Caption memory offers a way to carry context across an episode, but its benefit depends on what the model retains and how it uses that information. Converting video frames into text descriptions may also lose visual details needed to identify clinical issues. Future models need to retain relevant details from earlier scenes and use them to assess later events. MedSceneBench evaluates whether models identify the annotated issues and explain them correctly, rather than how much text they generate or how many issues they report.

Supplying the annotated clip did not resolve these difficulties. Even with the relevant scene given, most models classified the issue categories no better than random answers. The difficulty therefore lies not only in searching a long episode but also in judging which clinical concern a scene shows.

## Conclusion

MedSceneBench provides clinician-annotated hospital narratives for testing clinical issue discovery, temporal localization, and explanation. All 10 open models completed Direct-Full and five completed Caption-Memory-20m. Even across these runs, the models matched only 52 of 795 annotated issues, and few matched explanations identified the clinician's core concern. The benchmark supports evaluation of models that connect events across an episode and explains the clinical significance of the care shown.

## Acknowledgements

We thank Dr. Hayas Koya and Dr. Zaher Oueida, quality Improvement and patient safety officers in the Division of Hospital Medicine at SUNY Upstate Medical University, Syracuse, NY for helping ensure the quality of these annotations.

## Disclosures

The authors used AI-assisted technologies, the large language model tools Claude (Anthropic) and Codex (OpenAI), to help draft and edit the text of this manuscript, brainstorm and automate the research workflow but the authors were always in the loop. The authors reviewed and revised all the output produced with these tools and take full responsibility for the content of this work.

## References

1. Chaiyachati KH, Shea JA, Asch DA, et al. Assessment of inpatient time allocation among first-year internal medicine residents using time-motion observations. JAMA Intern Med 2019;179:760-767. DOI: 10.1001/jamainternmed.2019.0095.
2. Müskens JLJM, Kool RB, van Dulmen SA, Westert GP. Overuse of diagnostic testing in healthcare: a systematic review. BMJ Qual Saf 2022;31:54-63. DOI: 10.1136/bmjqs-2020-012576.
3. Glaser J, Nouri S, Fernandez A, et al. Interventions to improve patient comprehension in informed consent for medical and surgical procedures: an updated systematic review. Med Decis Making 2020;40:119-143. DOI: 10.1177/0272989X19896348.
4. Beltran-Aroca CM, Girela-Lopez E, Collazo-Chao E, Montero-Pérez-Barquero M, Muñoz-Villanueva MC. Confidentiality breaches in clinical practice: what happens in hospitals? BMC Med Ethics 2016;17:52. DOI: 10.1186/s12910-016-0136-y.
5. Kulkarni PA, Greene MT, Saint S, et al. Comparing observed with self-reported hand hygiene adherence among bedside nurses and physicians at two hospitals in the United States. Am J Infect Control 2026;54:241-246. DOI: 10.1016/j.ajic.2025.10.032.
6. Diem SJ, Lantos JD, Tulsky JA. Cardiopulmonary resuscitation on television — miracles and misinformation. N Engl J Med 1996;334:1578-1582. DOI: 10.1056/NEJM199606133342406.
7. Zago D, Cautero P, Scarpis E, et al. TV medical dramas: assessing the portrayal of public health in primetime. Front Public Health 2024;12:1432528. DOI: 10.3389/fpubh.2024.1432528.

8. Polk AO, DeMott SE, Mohorn PL, et al. Are medication orders in fictional medical-themed television shows accurate? A retrospective, observational study. Br J Med Med Res 2016;17:1-9. DOI: 10.9734/BJMMR/2016/28408.

9. Schmoll E, Seifert R. Pharmacological content in the television series "House MD": analysis from a German pharmacologist's perspective. Naunyn Schmiedebergs Arch Pharmacol 2026;399:10615-10641. DOI: 10.1007/s00210-026-04995-6.

10. Guest M, Johnson B, Pistono M, Kruser JM, Vasilevskis EE, Golden BP. Depiction of delirium in contemporary medical television dramas: a descriptive content analysis. J Am Geriatr Soc 2026;74:3108-3110. DOI: 10.1111/jgs.70565.

11. Cambra-Badii I, Pinar A, Baños JE. The Good Doctor and bioethical principles: a content analysis. Educ Med 2021;22:84-88. DOI: 10.1016/j.edumed.2019.12.006.

12. Grinstead E, Johnson A, Decker B, et al. Analysis of social determinants of health in medical dramas to enhance emergency medicine training. Am J Emerg Med 2026;110:153-157. DOI: 10.1016/j.ajem.2026.08.019.

13. Jackson KC, Miller KA, Hill SM, Johnson SS, Reimanis LA, Lofgren ET. Charting is never exciting: measuring physician time use in medical dramas. March 7, 2024 (https://doi.org/10.1101/2024.03.01.24303412). Preprint.

14. Czarny MJ, Faden RR, Sugarman J. Bioethics and professionalism in popular television medical dramas. J Med Ethics 2010;36:203-206. DOI: 10.1136/jme.2009.033621.

15. Hinkelbein J, Spelten O, Marks J, Hellmich M, Böttiger BW, Wetsch WA. An assessment of resuscitation quality in the television drama Emergency Room: guideline non-compliance and low-quality cardiopulmonary resuscitation lead to a favorable outcome? Resuscitation 2014;85:1106-1110. DOI: 10.1016/j.resuscitation.2014.05.009.

16. Portanova J, Irvine K, Yi JY, Enguidanos S. It isn't like this on TV: revisiting CPR survival rates depicted on popular TV shows. Resuscitation 2015;96:148-150. DOI: 10.1016/j.resuscitation.2015.08.002.

17. Hoffman BL, Hoffman R, Wessel CB, Shensa A, Woods MS, Primack BA. Use of fictional medical television in health sciences education: a systematic review. Adv Health Sci Educ Theory Pract 2018;23:201-216. DOI: 10.1007/s10459-017-9754-5.

18. Cambra-Badii I, Gomar-Sancho C, Mastandrea PB, et al. Cinemeducation to teach patient safety: an experience in medical students. Humanit Soc Sci Commun 2024;11:561. DOI: 10.1057/s41599-024-03054-w.

19. Lei J, Yu L, Bansal M, Berg T. TVQA: localized, compositional video question answering. In: Proceedings of the 2018 Conference on Empirical Methods in Natural Language Processing. Kerrville, TX: Association for Computational Linguistics, 2018:1369-1379. DOI: 10.18653/v1/D18-1167.

20. Fu C, Dai Y, Luo Y, et al. Video-MME: the first-ever comprehensive evaluation benchmark of multi-modal LLMs in video analysis. In: Proceedings of the 2025 IEEE/CVF Conference on Computer Vision and Pattern Recognition. Piscataway, NJ: IEEE, 2025:24108-24118. DOI: 10.1109/CVPR52734.2025.02245.

21. Wu H, Li D, Chen B, Li J. LongVideoBench: a benchmark for long-context interleaved video-language understanding. In: Advances in Neural Information Processing Systems 37 (Datasets and Benchmarks Track). Red Hook, NY: Curran Associates, 2024:28828-28857. DOI: 10.52202/079017-0907.
22. Zhou J, Shu Y, Zhao B, et al. MLVU: benchmarking multi-task long video understanding. In: Proceedings of the 2025 IEEE/CVF Conference on Computer Vision and Pattern Recognition. Piscataway, NJ: IEEE, 2025:13691-13701. DOI: 10.1109/CVPR52734.2025.01278.
23. Gupta D, Attal K, Demner-Fushman D. A dataset for medical instructional video classification and question answering. Sci Data 2023;10:158. DOI: 10.1038/s41597-023-02036-y.
24. Wang G, Wang J, Mo W, et al. SurgVidLM: towards multi-grained video understanding with large language model in robot-assisted surgery. In: Proceedings of the 2026 IEEE International Conference on Robotics and Automation. Piscataway, NJ: IEEE, 2026:13177-13184. DOI: 10.1109/ICRA57385.2026.11697585.
25. Su Y, Choudhuri A, Gao Z, et al. MedGRPO: multi-task reinforcement learning for heterogeneous medical video understanding. In: Proceedings of the 2026 IEEE/CVF Conference on Computer Vision and Pattern Recognition. Piscataway, NJ: IEEE, 2026:2788-2798.
26. Özsoy E, Pellegrini C, Czempiel T, et al. MM-OR: a large multimodal operating room dataset for semantic understanding of high-intensity surgical environments. In: Proceedings of the 2025 IEEE/CVF Conference on Computer Vision and Pattern Recognition. Piscataway, NJ: IEEE, 2025:19378-19389. DOI: 10.1109/CVPR52734.2025.01805.
27. Özsoy E, Pellegrini C, Bani-Harouni D, Yuan K, Keicher M, Navab N. Specialized foundation models for intelligent operating rooms. NPJ Digit Med 2026;9:362. DOI: 10.1038/s41746-026-02631-4.
28. Johnson KB, Alasaly B, Jang KJ, Eaton E, Mopidevi S, Koppel R. Observer: creation of a novel multimodal dataset for outpatient care research. J Am Med Inform Assoc 2026;33:424-433. DOI: 10.1093/jamia/ocaf182.
29. Du B, Liu B, Yu Y, et al. MedHorizon: towards long-context medical video understanding in the wild. In: Proceedings of the 2026 Conference on Neural Information Processing Systems (Evaluations and Datasets Track). 2026 (https://neurips.cc/virtual/2026/poster/139391).
30. Wang Y, Gao S, Jiang S, Hu Z, Liu Z. MedStreamBench: a time-aware benchmark for streaming and proactive medical video understanding. July 2, 2026 (https://arxiv.org/abs/2607.01751). Preprint.
31. Kang S, Holcomb M, Hein D, Shakur AH, Dalton T, Jamieson A. Physical examination identification in medical education videos: zero-shot multimodal AI with temporal sequence optimization study. JMIR AI 2025;4:e76586. DOI: 10.2196/76586.
32. Geng T, Zhang J, Wang Q, Wang T, Duan J, Zheng F. LongVALE: vision-audio-language-event benchmark towards time-aware omni-modal perception of long videos. In: Proceedings of the 2025 IEEE/CVF Conference on Computer Vision and Pattern Recognition. Piscataway, NJ: IEEE, 2025:18959-18969. DOI: 10.1109/CVPR52734.2025.01766.
33. Asgarov A, Narasimhan K, Sarker NH, et al. NEST: narrative event structures in time for long video understanding. In: Proceedings of the 2026 Conference on Empirical Methods in Natural

Language Processing. Kerrville, TX: Association for Computational Linguistics, 2026:57094-57129. DOI: 10.18653/2026.emnlp-main.2512.

34. Sushil M, Kennedy VE, Mandair D, Miao BY, Zack T, Butte AJ. CORAL: expert-curated oncology reports to advance language model inference. NEJM AI 2024;1(4). DOI: 10.1056/AIdbp2300110.

35. Jiang Y, Black KC, Geng G, et al. MedAgentBench: a virtual EHR environment to benchmark medical LLM agents. NEJM AI 2025;2(9). DOI: 10.1056/AIdbp2500144.

36. Liu Y, Ma Z, Qi Z, Wu Y, Shan Y, Chen CW. E.T. Bench: towards open-ended event-level video-language understanding. In: Advances in Neural Information Processing Systems 37 (Datasets and Benchmarks Track). Red Hook, NY: Curran Associates, 2024:32076-32110. DOI: 10.52202/079017-1009.

37. Grauman K, Westbury A, Byrne E, et al. Ego4D: around the world in 3,000 hours of egocentric video. In: Proceedings of the 2022 IEEE/CVF Conference on Computer Vision and Pattern Recognition. Piscataway, NJ: IEEE, 2022:18973-18990. DOI: 10.1109/CVPR52688.2022.01842.

38. Zhang T, Kishore V, Wu F, Weinberger KQ, Artzi Y. BERTScore: evaluating text generation with BERT. In: Proceedings of the 8th International Conference on Learning Representations. 2020 (https://openreview.net/forum?id=SkeHuCVFDr).

39. Fang X, Mao K, Duan H, et al. MMBench-Video: a long-form multi-shot benchmark for holistic video understanding. In: Advances in Neural Information Processing Systems 37 (Datasets and Benchmarks Track). Red Hook, NY: Curran Associates, 2024:89098-89124. DOI: 10.52202/079017-2827.

40. Li Y, Liu J, Zhang T, et al. Baichuan-Omni-1.5 technical report. January 26, 2025 (https://arxiv.org/abs/2501.15368). Preprint.

41. Cui J, Xu B, Wang C, et al. MiniCPM-o 4.5: towards real-time full-duplex omni-modal interaction. April 30, 2026 (https://arxiv.org/abs/2604.27393). Preprint.

42. Bai S, Cai Y, Chen R, et al. Qwen3-VL technical report. November 26, 2025 (https://arxiv.org/abs/2511.21631). Preprint.

43. Qwen Team. Qwen3.8-Max: a new bar for coding and cowork. August 2026 (https://qwen.ai/blog?id=qwen3.8).

44. Yu T, Wang Z, Wang C, et al. MiniCPM-V 4.5: cooking efficient MLLMs via architecture, data, and training recipe. In: Proceedings of the 2026 IEEE/CVF Conference on Computer Vision and Pattern Recognition. Piscataway, NJ: IEEE, 2026:11704-11715.

45. V Team, Hong W, Yu W, et al. GLM-4.5V and GLM-4.1V-Thinking: towards versatile multimodal reasoning with scalable reinforcement learning. July 1, 2025 (https://arxiv.org/abs/2507.01006). Preprint.

46. Kimi Team, Du A, Yin B, et al. Kimi-VL technical report. April 10, 2025 (https://arxiv.org/abs/2504.07491). Preprint.

47. Wang W, Gao Z, Gu L, et al. InternVL3.5: advancing open-source multimodal models in versatility, reasoning, and efficiency. August 25, 2025 (https://arxiv.org/abs/2508.18265). Preprint.

48. Zhang B, Li K, Cheng Z, et al. VideoLLaMA 3: frontier multimodal foundation models for image and video understanding. January 22, 2025 (https://arxiv.org/abs/2501.13106). Preprint.

49. Xu J, Guo Z, Hu H, et al. Qwen3-Omni technical report. September 22, 2025 (https://arxiv.org/abs/2509.17765). Preprint.